\documentclass[10pt,twocolumn,letterpaper]{article}

\usepackage{cvpr}              

\usepackage{graphicx}
\usepackage{amsmath}
\usepackage{amssymb,tipa}
\usepackage{booktabs}
\usepackage{multirow}
\usepackage{cite}
\usepackage{color}
 \usepackage{bbding}
 \usepackage{bm}

\usepackage[pagebackref,breaklinks,colorlinks]{hyperref}

\newcommand\blfootnote[1]{%
  \begingroup
  \renewcommand\thefootnote{}\footnote{#1}%
  \addtocounter{footnote}{-1}%
  \endgroup
}

\usepackage[capitalize]{cleveref}
\crefname{section}{Sec.}{Secs.}
\Crefname{section}{Section}{Sections}
\Crefname{table}{Table}{Tables}
\crefname{table}{Tab.}{Tabs.}
\newcommand{\tabincell}[2]{\begin{tabular}{@{}#1@{}}#2\end{tabular}}

\def\confName{CVPR}
\def\confYear{2023}

\begin{document}

\title{You Can Ground Earlier than See: An Effective and Efficient Pipeline for Temporal Sentence Grounding in Compressed Videos}
\author{
Xiang Fang\textsuperscript{1*} \ \
Daizong Liu\textsuperscript{2*} \ \
Pan Zhou\textsuperscript{1$\dagger$} \ \
Guoshun Nan\textsuperscript{3} \ \
\\
\textsuperscript{1}The Hubei Engineering Research Center on Big Data Security, School of Cyber Science \\ and Engineering,  Huazhong University of Science and Technology \\
\textsuperscript{2}Peking University \ \
\textsuperscript{3}Beijing University of Posts and Telecommunications \\
{\tt\small xfang9508@gmail.com  \quad dzliu@stu.pku.edu.cn \quad panzhou@hust.edu.cn \quad nanguo2021@bupt.edu.cn}
}

\maketitle

\begin{abstract}
Given an untrimmed video, temporal sentence grounding (TSG) aims to locate a target moment semantically according to a sentence query. 
Although previous respectable  works have made decent success, they only focus on high-level visual features extracted from the consecutive decoded frames and fail to handle the compressed videos for query modelling, suffering from insufficient representation capability and significant computational complexity during training and testing. 
In this paper, we pose a new setting, compressed-domain TSG, which directly utilizes compressed videos rather than fully-decompressed frames as the visual input.
To handle the raw video bit-stream input, we propose a novel Three-branch Compressed-domain Spatial-temporal Fusion (TCSF) framework,
which extracts and aggregates three kinds of low-level visual features (I-frame, motion vector  and residual features) for effective and efficient grounding. 
Particularly, instead of encoding the whole decoded frames like previous works, we capture the appearance representation by only learning the I-frame feature to reduce delay or latency. 
Besides, we explore the motion information not only by learning the motion vector feature, but also by exploring the relations of neighboring frames via the residual feature.
In this way, a three-branch spatial-temporal attention layer with an adaptive motion-appearance fusion module is further designed to extract and aggregate both appearance and motion information
for the final grounding.
Experiments on three challenging datasets 
shows that our TCSF achieves better performance than other state-of-the-art methods with lower complexity.
\end{abstract}
\vspace{-36pt}

\blfootnote{
\textsuperscript{$*$}Equal contributions. ~~~~\textsuperscript{$\dagger$}Corresponding author.}

\section{Introduction}
\begin{figure}[t!]
\centering
\includegraphics[width=0.48\textwidth]{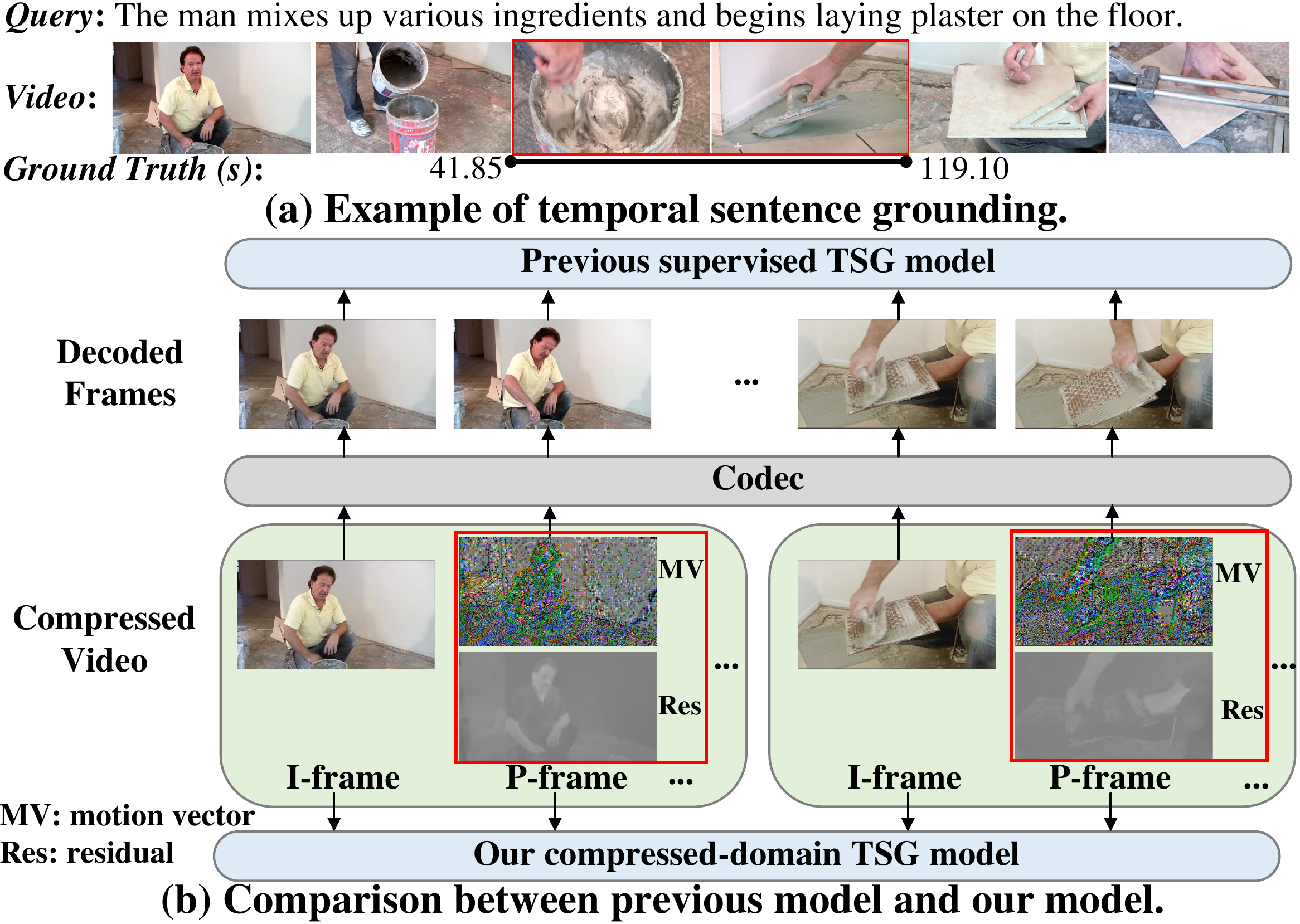}
\vspace{-8pt}
\caption{(a) Example of the temporal sentence grounding (TSG). (b) Comparison between previous supervised TSG models and our compressed-domain TSG model. Previous models first decode the video into consecutive frames and then feed them into their networks, while our compressed-domain model directly leverages the compressed video as the visual input.}
\vspace{-14pt}
\label{fig:intro}
\end{figure}
As a significant yet challenging computer vision task, temporal sentence grounding (TSG) has drawn increasing attention due to its various applications, such as video understanding \cite{zheng2022weakly,yang2022tubedetr,jiang2022semi,liu2023exploring,wang2025taylor,fang2026towardsicml,kuai2026dynamic,wang2025point,fang2025your,zhang2025monoattack,fang2023hierarchical,liu2024towards,yang2025eood,fang2022multi,fang2026cogniVerse,lei2025exploring,fang2020double,wang2025dypolyseg,fang2025hierarchical,yan2026fit,fang2025adaptive,wang2026topadapter,cai2025imperceptible,fang2026slap,wang2026reasoning,fang2026immuno,wang2026biologically,fang2026disentangling,wang2025reducing,fang2026advancing,fang2026unveiling,wang2026from,liu2023conditional,liu2026attacking,fang2026rethinking,wang2025seeing,fang2026towards,fang2025multi,fang2024fewer,liu2024pandora,fang2024multi,fang2025turing,fang2024not,liu2023hypotheses,fang2024rethinking,liu2024unsupervised,fang2023annotations,xiong2024rethinking,fang2021unbalanced,wang2025prototype,zhang2025manipulating,fang2026align,tang2024reparameterization,fang2025adaptivetai,tang2025simplification,fang2021animc,cai2026towards,fang2020v} and temporal action localization \cite{wang2022negative,su2021stvgbert}. Given a long untrimmed video, the TSG task aims to locate the specific start and end timestamps of a video segment with an activity that semantically corresponds to a given sentence query. As shown in Figure \ref{fig:intro}(a), most of video contents are query-irrelevant, where only a short video segment matches the query. It is substantially more challenging since a well-designed method needs to not only model the complex multi-modal interaction among video and query, but also capture complicated context information for cross-modal semantics alignment.

By treating a video as a sequence of independent frames, most TSG methods \cite{zeng2021multi,tang2021frame,liu2022exploring1,zhang2021multi,liu2021adaptive,liu2021progressively,liu2022memory,cao2020strong,Liu_2021_CVPR,lei2020tvr,zhao2021cascaded,liu2020jointly,liu2022few,liu2022reducing,liu2022skimming} refer to the fully-supervised setting, where each frame is firstly fully decompressed from a video bit-stream and then manually annotated as query-relevant or query-irrelevant. Despite the decent progress on the grounding performance, these data-hungry methods severely rely on the fully decompression and numerous annotations, which are significantly labor-intensive and time-consuming to obtain from real-word applications. To alleviate this dense reliance to a certain extent, some weakly-supervised works \cite{ZhangLZZH20,2021LoGAN,ChenMLW19,DuanHGW0H18,MithunPR19,MaYKLKY20,LinZZWL20,song2020weakly,zhang2020counterfactual} are proposed to only leverage the coarse-grained video-query annotations instead of the fine-grained frame-query annotations. Unfortunately, this weak supervision still requires the fully-decompressed video for visual feature extraction. 

Based on the above observation, in this paper, we make the first attempt to explore if an effective and efficient TSG model can be learned without the limitation of the fully decompressed video input. 
Considering that the real-world video always stored and transmitted in a compressed data format, we explore a more practical but challenging task: compressed-domain TSG, which directly leverages the compressed video instead of obtaining consecutive decoded frames as visual input for grounding. 
As shown in the Figure \ref{fig:intro}(b), a compressed video is generally parsed by a stream of Group of successive Pictures (GOPs) and each GOP starts with one intra-frame (I-frame) followed by a variable number of predictive frames (P-frames) \cite{xu2022accelerating,li2022end}. Specifically, the I-frame contains complete RGB information of a video frame, while each P-frame contains a motion vector and a residual. The motion vectors store 2D displacements between I-frame and its neighbor frames, and the residuals store the RGB differences between I-frame and its reconstructed frame calculated by Motion Vectors in the P-frames after motion compensation. 
The I-frame can be decoded itself, while these P-frames only store the changes from the previous I-frame by motion vectors and residuals.

Given the compressed video, our main challenge is how to effectively and efficiently extract contextual visual features from the above three low-level visual information for query alignment. Existing TSG works \cite{zeng2021multi,tang2021frame,liu2022exploring1,zhang2021multi,cao2020strong,Liu_2021_CVPR,lei2020tvr,zhao2021cascaded} cannot be applied directly to the compressed video because their video features (\emph{e.g.,} C3D and I3D) can only be extracted if all complete video frames are available after decompression. Moreover, decompressing all the frames will significantly increase computational complexity for feature extraction, leading to extra latency and extensive storage. 


To address this challenging task, we propose the first and novel approach for compressed-domain TSG, called Three-branch Compressed-domain Spatial-temporal
Fusion (TCSF). 
Given a group of successive picture (GOP) in a compressed video, we first extract the visual features from each I-frame to represent the appearance at its timestamp, and then extract the features of its P-frames to capture the motion information near the I-frame.
In this way, we can model the activity content with above simple I-frame and P-frames instead of using their corresponding consecutive decoded frames.
Specifically, we design  a spatial attention and a temporal attention to integrate the appearance and motion features for activity modelling. 
To adaptively handle different fast-motion (P-frame guided) or slow-motion (I-frame guided) cases, we further design an adaptive appearance and motion fusion module to integrate the appearance and motion information by learning a balanced weight through a residual module. Finally, a query-guided multi-modal fusion is exploited to integrate the visual and textual features for final grounding.

Our contributions are summarized as follows:
\begin{itemize}
    \item We propose a brand-new and challenging task: compressed-domain TSG, which aims to directly leverage the compressed video for TSG. To our best knowledge, we make the first attempt to locate the target segment in the compressed video.
    \item We present a novel pipeline for compressed-domain TSG, which can efficiently and effectively  integrate both appearance and motion information from the low-level visual information in the compressed video.
    \item Extensive experiments on three challenging datasets (ActivityNet Captions, Charades-STA and TACoS) validate the effectiveness and efficiency of our TCSF.
\end{itemize}

\section{Related Works}
\noindent \textbf{Temporal sentence grounding.} Most existing TSG methods are under the fully-supervised setting, where all video-query pairs and precise segment boundaries are manually annotated based on the fully-decompressed video.
These methods can be divided into two categories:
1) Proposal-based methods \cite{anne2017localizing,chen2018temporally,zhang2019cross,yuan2019semantic,zhang2019learning}: They first pre-define multiple segment proposals and then align these proposals with the query for cross-modal semantic matching based on the similarity. Finally, the best proposal with the highest similarity score is selected as the predicted segment. 
Although achieving decent results, these proposal-based methods severely rely on the quality of the segment proposals and are time-consuming.
2) Proposal-free methods \cite{chenrethinking,yuan2019find,mun2020local,zhang2020span,liu2022unsupervised}: They directly regress the start and end boundary frames of the target segment or predict boundary probabilities frame-wisely. 
Compared with the proposal-based methods, proposal-free methods are more efficient.
To alleviate the reliance to a certain extent, some state-of-the-art turn to the weakly-supervised setting \cite{ZhangLZZH20,2021LoGAN,ChenMLW19,DuanHGW0H18,MithunPR19,MaYKLKY20,LinZZWL20,song2020weakly,zhang2020counterfactual}, where only video-query pairs are annotated without precise segment boundaries in the fully-decompressed video.

In real-world computer vision tasks, we always collect the compressed video, rather than decompressed consecutive frames. In this paper, we present a brand-new practical yet challenging setting for TSG task, called compressed-domain TSL, with merely compressed video rather than a decompressed frame sequence.

\noindent \textbf{Video compression.} As a fundamental computer vision task, video compression \cite{xu2021detection,lu2022learning,lauzon1996performance,lin2009versatile,lee2006adaptive,wiegand2003overview,schwarz2007overview} divides a video into a group of pictures (GOP), where each frame is coded as an I-, P-, and B- frame. An I-frame is the first frame of the GOP to maintain full RGB pixels as an anchor. The subsequent P-and B-frames are then coded using a block-based motion vector with temporal prediction. The prediction is conducted by searching the closest matching block of a previously coded frame as a reference frame. A vector of the current block to the reference block is determined as a motion vector. Since the current block and the matching block are often different, the transformed residual is used to denote the difference.


Compared with other deep features (\textit{e.g.,} optical flow \cite{ilg2017flownet}) widely used in the TSG task, the compressed-domain features (MVs and residual) have the following advantages:
1) Lower computational costs. The compressed-domain features can be obtained during decoding, while other deep features need to decompress  the compressed video and encode the video by a pretrained heavy-weight model (C3D \cite{tran2015learning} or I3D \cite{carreira2017quo}). The compressed-domain features only 
even require partial-frame reconstruction by entropy decoding \cite{zhu2007pop}, inverse transform and quantization \cite{lee2016compressed}, and motion-compensation \cite{divakaran2000video}. In entropy decoding, the most time-consuming process is skipping the motion-compensation \cite{shen2005submacroblock}, whose computational complexity is much smaller than that of other deep features. 2) No delay or dependency. The compressed-domain features can be instantly obtained. When we large-scale datasets, the advantages are more obvious. 



\section{Proposed Method}
\begin{figure*}[t!]
\centering
\includegraphics[width=\textwidth]{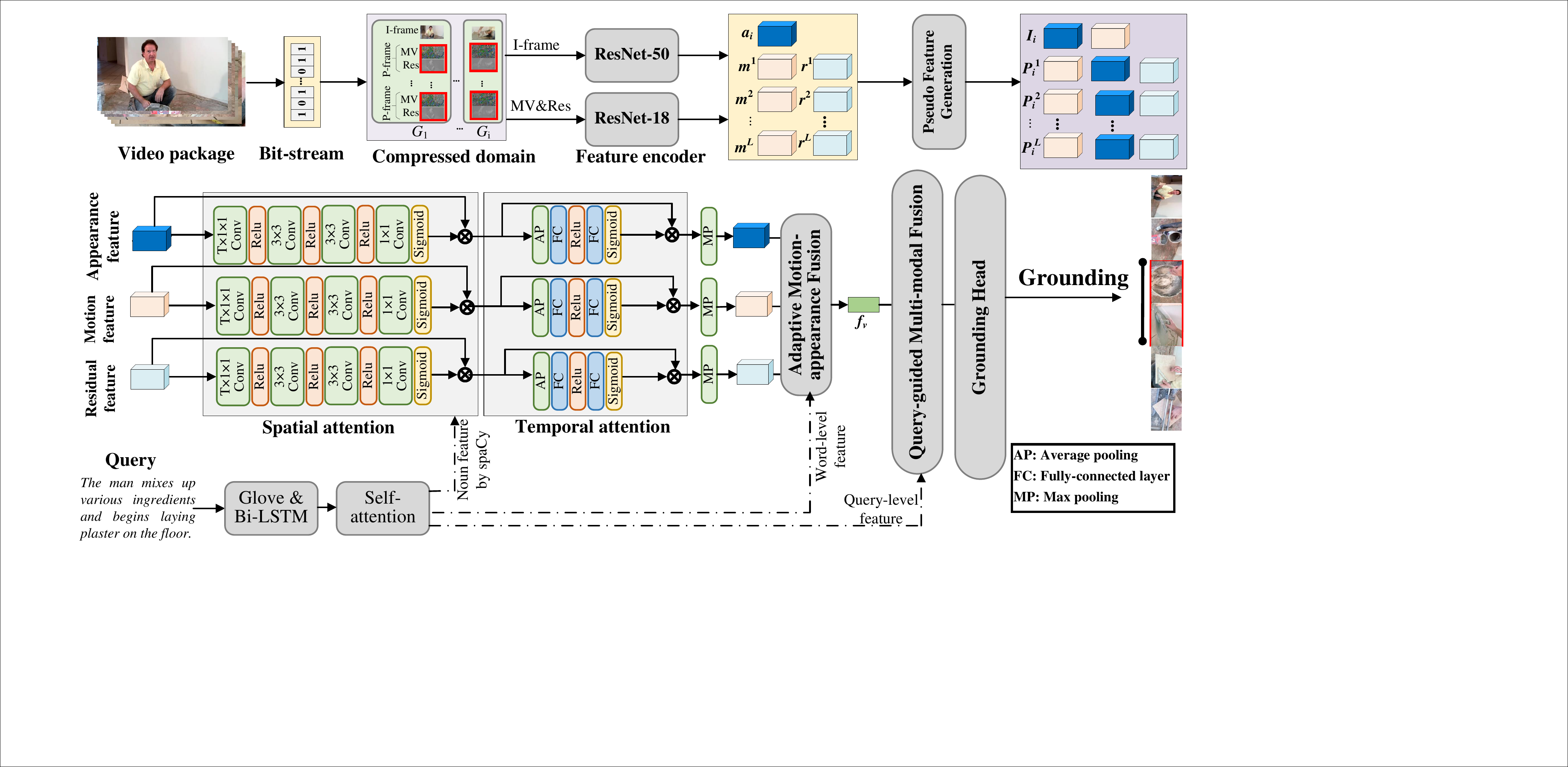}
\vspace{-20pt}
\caption{Overview of the proposed architecture. Firstly, we  leverage the entropy decoding approach to obtain the compressed video, \textit{i.e.}, I-frames and P-frames (containing motion vectors and residuals). Then, we enrich their information with pseudo features, and develop a three-branch spatial-temporal attention to model the query-related activity content. After that, we fuse the appearance and motion contexts, and integrate them with the query features for learning the joint multi-modal representations. At last, we feed the multi-modal features into the grounding head to predict the segment.}
\vspace{-10pt}
\label{fig:pipeline}
\end{figure*}
\subsection{Overview}
\noindent \textbf{Problem statement.}
Given a video bit-stream $\mathcal{V}$ with $T$ frames, the temporal sentence grounding (TSG) task aims to localize the precise boundary $(\tau_s, \tau_e)$ of a specific segment semantically corresponding to  a given query $\mathcal{Q}=\{q_j\}_{j=1}^M$, where $q_j$ denotes the $j$-th word, $M$ denotes the word number, $\tau_s$ and $\tau_e$ denote the start and end timestamps of the specific segment.
In our compressed-domain TSG setting, we do not feed the decompressed frames video as input. Instead, we partially decode the video bit-stream at a low cost to extract the compressed video, which includes $N$  group of pictures (GoPs). Each GoP $G_i$ contains one reference I-frame $I_i\in \mathbb{R}^{\mathcal{H}\times\mathcal{W}\times 3}$ followed by $L$ number of P-frames $\{P_i^l\}_{l=1}^L$.
Each $P_i^l$ consists of a motion vector $M_i^l\in \mathbb{R}^{\mathcal{H}\times \mathcal{W} \times 2}$ and a residual $R_i^l\in \mathbb{R}^{\mathcal{H}\times \mathcal{W} \times 3}$, which can be extracted nearly cost-free from $\mathcal{V}$.
For convenience, we assume that all GOPs contain  the same number of P-frames. Thus, $T=N\times(L+1)$. 
The video bit-stream can be represented as $\mathcal{V}=\{I_i, P_i^1, P_i^2, \cdots, P_i^{L}\}_{i=1}^N$, 
where $i$ denotes the $i$-th GOP.
Here, the I-frame contains complete RGB information of a video frame and can be decoded itself, while these P-frames only store the changes from the previous I-frame by motion vectors and residuals. The motion vectors store 2D displacements of the most similar patches between I-frame and the target frame, and the residuals store pixel-wise differences to correct motion compensation errors. 
We use above three low-level information contained in compressed videos as our visual input.

\noindent \textbf{Pipeline.} 
Our pipeline is summarized in Figure~\ref{fig:pipeline}. 
Given a video bit-stream, we first utilize the entropy decoding approach \cite{wu2018compressed,wang2019fast} to generate a group of successive pictures (GOP), which consists of several I-frames with their related P-frames.
Then, we extract the visual appearance features 
from I-frames by a pre-trained ResNet-50 network, while a light-weight ResNet-18 network is used to extract the motion vector and residual features from P-frames. 
After that, we enrich these partial appearance and motion information with pseudo features to make the complete comprehension of the full video.
A spatial-temporal attention module is further introduced to better model the activity content based on the motion-appearance contexts.  
Next, we design an adaptive appearance and motion fusion module to selectively integrate the attentive appearance and motion information guided by the residual information. Finally, we design a query-guided multi-modal fusion module to integrate the visual and textual features for final grounding.

\subsection{Multi-Modal Encoding}
\noindent \textbf{Query encoder.}  Following \cite{gao2017tall}, we first employ the Glove network \cite{pennington2014glove} to embed each word into a dense vector.
Then, a Bi-GRU network \cite{chung2014empirical} and a multi-head self-attention module \cite{vaswani2017attention} are used to further integrate the sequential textual representations. Thus, final word-level features is denote as $Q=\{q_j\}_{j=1}^M\in \mathbb{R}^{M \times d}$, where $d$ is the feature dimension. By concatenating the outputs of the last hidden unit in Bi-GRU with a further linear projection, we can obtain the sentence-level feature as $q_{global} \in \mathbb{R}^{d}$.


\noindent \textbf{I-frame encoder.} Following \cite{marpe2010video,li2020efficient},
if the $\{t\}_{t=1^T}$-th frame is I-frame,
we use a pretrained ResNet-50 model \cite{he2016deep} to extract its appearance feature $a^t\in \mathbb{R}^{H\times W \times C}$, where $H$, $W$ and $C$ denotes dimensions of height, width, and channel. 

\noindent \textbf{P-frame encoder.}  Following \cite{shou2019dmc, wu2018compressed},
if the $\{t\}_{t=1^T}$-th frame is P-frame containing a motion vector $M^t$ and a residual $R^t$, we utilize a ResNet-18 network \cite{he2016deep} to extract the motion vector feature $m^t\in \mathbb{R}^{H\times W \times C}$ and the residual feature $r^t\in \mathbb{R}^{H\times W \times C}$.

\noindent \textbf{Pseudo feature generation.}
Since our compressed-domain TSG needs to locate the specific start and end frames of the target segment, we need to obtain the precise motion, compensation and appearance information of each frame for more accurate grounding. 
However, in the compressed video, we only have partially $N$-number I-frames of appearance and $(N\times L)$-number P-frames of motion and compensation, lacking enough full-frames (\textit{i.e.}, $T$-number frames) knowledge of the complete appearance-motion information. 
Thus, we tend to generate complementary pseudo features for the unseen frames of the video. For example, to warp the appearance feature from the current I-frame,
we can use $M^t$ to estimate the pseudo appearance feature $a^{t+1}$ in its adjacent frame (its next frame).
We can find that the pseudo feature generation approach exempts reconstructing each adjacent frame for feature extraction individually. 
We assume that the $t$-frame is I-frame.
For constructing the pseudo appearance features of its $n$-th adjacent P-frame, we utilize a block-based motion estimation as:
\small
\begin{align}\label{appear_generate}
a^{n+t}(s)=a^{n+t-1}(\delta M^{n+t-1}(s\delta)+s),
\end{align}\normalsize
where $a^{n+t}$ denotes the appearance feature of the $n+t$-th P-frame, $s$ is a spatial coordinate of features, and $\delta$ is used as a scaling factor.  By Eq. \eqref{appear_generate}, we can obtain the appearance information of each P-frame based on off-the-shelf I-frames.

Similarly, we will generate the motion information of each I-frame based on P-frames. Following \cite{feichtenhofer2019slowfast},  we  combine the temporal movement information of appearance features in these adjacent frames.
In the channel axis, we concatenate consecutive $n$ frames $[a^t;\cdots;a^{n+t}]$ as $V^t\in \mathbb{R}^{H \times W\times C\times n}$.
Setting $V_*^t=conv_{1\times 1}(V^t)$, we can get
\small
\begin{align}\label{motion_generate}
m^t=ReLU(V_*^t),
\end{align}\normalsize
where $m^t$ is the motion feature of $t$-th frame, ReLU is the ReLU function,  and $conv_{1\times 1}$ means $1\times 1$ convolution layer with stride 1, producing a channel dimension of feature $C\times n$ to $C$. Thus, for the $t$-th frame, 
 its appearance and motion features are $a^t$ and $m^t$, respectively.

\subsection{Three-branch Spatial-temporal Attention}
In the TSG task, most of regions within a frame are query-irrelevant, where only a few regions are query-relevant. 
To automatically learn the discriminative regions relevant to the query, we need to obtain the fine-grained local spatial context.
Besides, the temporal context is also important since we can correlate the region-attentive spatial information in time series for precisely modelling the activity.
Therefore, we exploit previous-encoded three low-level features (appearance, motion and residual features) to obtain such query-relevant temporal-spatial information by designing a three-branch temporal and spatial attention.

\noindent \textbf{Spatial attention.} We propose the spatial attention to guide the model put more focus on the query-related region of the low-level features. Specifically, in the TSG task, most spatial visual information is noun-relevant. We first utilize the NLP tool spaCy \cite{honnibal2017natural} to parse nouns from the given query. Then, we exploit these nouns to enhance three visual features (appearance, motion and residual features) via an attention mechanism for helping the model learn to pay more attention on the spatial information precisely. 
The details of spatial attention is shown in Figure \ref{fig:pipeline}, where we leverage the combination of two 2D convolutional layers with kernel size of $3\times 3$, two RelUs and a 2D convolutional layers with kernel size of $1\times 1$ to obtain the spatial attention map. 
Therefore, we can enhance the region-attentive appearance features $a^t$ into $a_*^t$. Similarly, we can also obtain the region-attentive motion feature $m_*^t$ and region-attentive residual features $r_*^t$.

\noindent \textbf{Temporal attention.}
After learning the region-aware spatial information, we further learn to capture their temporal relation to better model the query-relevant activity. 
Specifically, we choose $K$ consecutive frames (starting at the $t$-th frame) for extracting their temporal information via a newly proposed temporal attention.
Here, we take the temporal attention on appearance features for example.
For the appearance features, we first concatenate them as $\mathcal{A}=[a_*^t;\cdots;a'^{t+K-1}]$. To yield the temporal weights $w=[w^1,w^2,\cdots,w^K]\in\mathbb{R}^{K}$ on these consecutive frames, we first leverage a global average pooling along three dimensions $H\times W \times C$ to generate a temporal-wise statistics $S=[s^1,s^2,\cdots,s^K]\in\mathbb{R}^{K}$, where $s^t$ represents the whole temporal information of $w^t$. Then, we utilize the temporal attention module shown in Figure~\ref{fig:pipeline} to generate the temporal weights $w^t$ as:
\small
\begin{align}
w^t=\sigma(W_{F1} \circ ReLU(W_{F2}\circ S+b_2)+b_1),
\label{w_c}
\end{align}\normalsize
where $W_{FC_1}$ and $W_{FC_2}$ are the weights of two FC layers;  $b_1\in\mathbb{R}^{K}$ and $b_2\in\mathbb{R}^{K}$ are the biases of two FC layers; $\circ$ denotes the convolution operation. Therefore, the final output of the appearance branch is:
\small
\begin{align}\label{channel_out}
f_{\hat{a}}^t=w_ca_*^t.
\end{align}\normalsize
Similarly, we can obtain the final outputs of the MV and residual branches as: $f_{\hat{m}}^t$ and $f_{\hat{r}}^t$.

\begin{figure}[t!]
\centering
\includegraphics[width=0.4\textwidth]{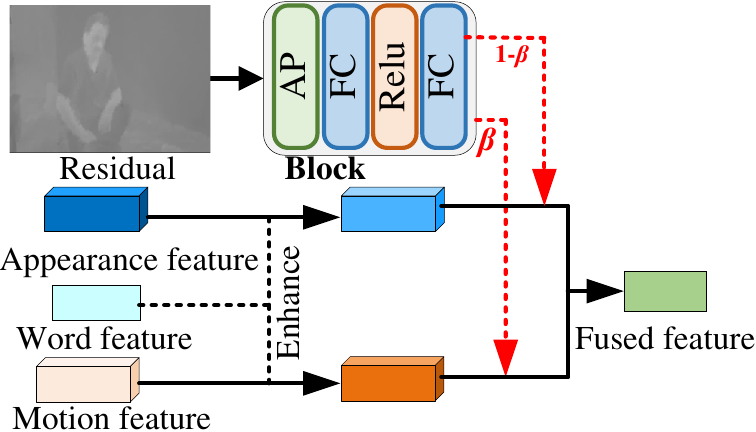}
\vspace{-4pt}
\caption{Framework of adaptive motion-appearance fusion.}
\label{fig:res}
\vspace{-12pt}
\end{figure}
\subsection{Adaptive Motion-Appearance Fusion} 
After obtaining the attentive motion and appearance information, we tend to aggregate them to infer the activity content.
Considering different videos may contain different abrupt temporal changes, we cannot equally fuse both the motion and appearance.
Specifically,
in the TSG task, a video with more abrupt temporal changes often corresponds to a related word. For example, a video corresponding to ``run'' often have more temporal changes than another video corresponding to ``walk''.
Therefore, we propose an adaptive strategy to fuse motion and appearance reasonably.
Specifically, we first enhance the appearance and motion features based on the query features. Then, we leverage the residual information to balance the enhanced appearance features and the enhanced motion features adaptively.

\noindent \textbf{Query-guided feature enhancement.}
We first utilize an attention mechanism to aggregate the word-level query features $\{q_j\}_{j=1}^M$ for each appearance feature $f_{\hat{a}}^t$ as:
\small
\begin{align}\label{attention_appearance}
X^t&=W^{\top} \text{tanh}(\bm{W}_1 f_{\hat{a}}^t + \bm{W}_2 \bm{q}_j + \bm{b}_0),\\
    A_a^t &= \frac{X^t}{\sum_sX^t(s)},
\end{align}\normalsize
where $A_a^t$ is the attention,
$\bm{W}_1$ and $\bm{W}_2$ are projection matrices, $\bm{b}_0$ is the bias vector, and the $W^{\top}$ is the row vector as in \cite{zhang2019cross}. Based on Eq. \eqref{attention_appearance}, we can obtain the query-enhanced appearance feature $f_a^t=f_{\hat{a}}^t\odot  A_a^t$, where $\odot$ denotes the operation of element-wise product. Similarly, we can obtain the query-enhanced motion feature $f_m^t$, which also semantically corresponds to the query.

\noindent \textbf{Residual-guided feature fusion.}
The residual features not only represent the temporal changes (\textit{i.e.}, motion context) among adjacent frames, but also denote the changes occur in RGB pixels (\textit{i.e.}, appearance context).
Thus, we utilize the residual feature as guidance to synchronize motion and appearance features by
a learnable Block (shown Figure \ref{fig:res}):
\small
\begin{align}\label{beta}
\beta^t=Block(f_{\hat{r}}^t),
\end{align}\normalsize
where $\beta^t\in[0,1]$ is a learnable balance, the block contains an average pooling, two fully-connected layers and a RelU network. If there are many abrupt temporal changes between different scenes, $\beta^t$ will approach 1. On the contrary, when there are few abrupt temporal changes, $\beta^t$ goes nearly to 0. At last, we fuse the motion and appearance information with this balanced weight as:
\small
\begin{align}\label{f_fuse}
f_{v}^t=\beta W_3f_a^t+(1-\beta)W_4f_m^t,
\end{align}\normalsize
where matrices $W_3$ and $W_4$ are learnable parameters.

\subsection{Multi-modal Fusion and Grounding Head}
After obtaining the motion-appearance enhanced visual feature, we further integrate it with the textual features as:
\small
\begin{align}
o=W_5\sum_{t=1}^{T}f_v^t+W_6\sum_{j=1}^Mq_j+W_7q_{global}.
\end{align}\normalsize
where $o$ is the fused feature, and $W_5$, $W_6$ and $W_7$ are learnable weight matrices.
Based on the multi-modal features $o$,  
we utilize two separate Multilayer Perceptron (MLP) layers to predict the start and end scores on each video clip as
\small
\begin{align}
\gamma = \text{softmax} (\text{MLP}_{1}(o)); 
(\tau_s, \tau_e) = \text{MLP}_{\text{reg}}(\sum_{t=1}^{T} \gamma^t o^t),
\label{eq:regression}
\end{align}\normalsize
where $\gamma \in \mathbb{R}^{T}$ is the attention weights for segments.
Following \cite{mun2020local}, we introduce 
the regression loss $\mathcal{L}_{\text{reg}}$ to learn the timestamp prediction as follows:
\small
\begin{align}
\mathcal{L}_{\text{reg}}= \mathcal{S}(\hat{\tau}_s-\tau_s) + \mathcal{S}(\hat{\tau}_e-\tau_e),
\label{eq:reg_loss}
\end{align}\normalsize
where $(\hat{\tau}_s, \hat{\tau}_e) \in [0,1]$ is the normalized ground-truth boundary, $\mathcal{S}(x)$ is the smooth $L_1$ function, which is defined as $|x|-0.5$ if $|x|\geq 1$ and $x^2/2$ otherwise.
We also introduce a confident loss $\mathcal{L}_{\text{guide}}$ to guide timestamp prediction:
\small
\begin{align}
\mathcal{L}_{\text{guide}} = -\frac{\sum_{t=1}^{{T}} \hat{\gamma}^t \log(\gamma^t)}{\sum_{t=1}^{{T}} \hat{\gamma}^t},
\label{eq:guide_loss}
\end{align}\normalsize
where $\hat{\gamma}^t=1$ if the $t$-th segment is located within the ground-truth boundary and $\hat{\gamma}^t=0$ otherwise.
By Eq. \eqref{eq:guide_loss}, we can obtain higher attention weights for the segments semantically relevant to the text query.

Therefore, the final loss function is formulated as:
\small
\begin{align}
\mathcal{L}_{final}=\mathcal{L}_{\text{reg}}+\alpha\mathcal{L}_{\text{guide}},
\end{align}\normalsize
where $\alpha$ is a hyper-parameter.

\noindent \textbf{Inference.} (i) Given a video bit-stream and a language query, we feed them into our TCSF to obtain the fused cross-modal feature $o$ in Eq. (9). (ii) We predict the moment boundary ($\tau_s$, $\tau_e$) by $o$ in Eq. (10) and the confidence score in Eq. (11). (iii) Based on predicted the coarse moment boundary and confidence scores, we generate several candidate moments, ``Top-n (R@n)'' candidates will be selected with non-maximum suppression.
\section{Experiment}
\begin{table}[t!]
\small
    \centering
    \caption{Effectiveness comparison for temporal sentence grounding on ActivityNet Captions dataset under official train/test splits.}
     \vspace{-11pt}
    \setlength{\tabcolsep}{1.5mm}{\begin{tabular}{c|c|cccc}
    \hline
    \multirow{2}*{Method} & \multirow{2}*{Type} & R@1, & R@1, & R@5, & R@5, \\ 
    ~ & ~ & IoU=0.3 & IoU=0.5 & IoU=0.3 & IoU=0.5\\ \hline 
CTRL \cite{gao2017tall}& FS & - & 29.01 & - & 59.17\\
2D-TAN \cite{zhang2019learning}& FS & 59.45 & 44.51 & 85.53 & 77.13 \\
 DRN \cite{zeng2020dense}& FS & - & 45.45 & - & 77.97\\
RaNet \cite{gao2021relation}&FS&-&45.59&-&75.93\\
 MIGCN \cite{zhang2021multi}& FS&-&48.02&-&78.02\\
MMN \cite{wang2022negative}&FS&65.05&48.59&87.25&79.50\\
\hline
ICVC \cite{chen2022explore}& WS & 46.62 & 29.52 & 80.92 & 66.61 \\
LCNet  \cite{yang2021local}& WS & 48.49 & 26.33 & 82.51 & 62.66\\
VCA \cite{wang2021visual}& WS & 50.45 & 31.00 & 71.79 & 53.83 \\
WSTAN \cite{wang2021weakly}& WS & 52.45 & 30.01 & 79.38 & 63.42\\
CNM \cite{zheng2022weakly}&WS&55.68&33.33&-&-\\
\hline\hline
\textbf{Our TCSF} & \textbf{CD} & \textbf{66.87}&\textbf{48.38}&\textbf{88.75}&\textbf{80.24}  \\\hline
 \end{tabular}}
  \vspace{-5pt}
    \label{tab:ActivityNet}
\end{table}
\subsection{Datasets}
\noindent \textbf{ActivityNet Captions.}
Built from ActivityNet v1.3 dataset \cite{caba2015activitynet} for the dense video captioning task, ActivityNet Captions contains 20k YouTube videos and 100k language queries. On average, a video are 2 minutes and a query has about 13.5 words.  Following the public split \cite{gao2017tall}, we use 37421, 17505, and 17031 video-query pairs for training, validation and testing.

\noindent \textbf{Charades-STA.}
Built upon the Charades  dataset \cite{sigurdsson2016hollywood,gao2017tall}, Charades-STA contains 16128 video-sentence pairs. Folowing \cite{gao2017tall}, we utilize 12408 pairs  for training and the others  for testing. The average video length is 0.5 minutes. The language annotations are generated by sentence decomposition and keyword matching with manual check.

\noindent \textbf{TACoS.}
Collected from the cooking scene by \cite{regneri2013grounding}, TACoS is employed for the video grounding and dense video captioning tasks. The dataset consists of 127 videos, whose average length is 4.8 minutes. Following the same split of \cite{gao2017tall}, we leverage 10146, 4589, and 4083 video-query pairs for training, validation, and testing respectively.

\subsection{Experimental Settings}
\noindent \textbf{Evaluation metric.}
Following \cite{gao2017tall,liu2018attentive,zhang2020span}, we evaluate the grounding performance by ``R@n, IoU=m'', which means the percentage of  queries having at least one result whose Intersection over Union (IoU) with ground truth is larger than m. In our experiments, we use $n \in \{1,5\}$ for all datasets, $m \in \{0.5,0.7\}$ for ActivityNet Captions and Charades-STA, $m \in \{0.3,0.5\}$ for TACoS.

\begin{table}[t!]
\small
    \centering
    \caption{Performance comparison for temporal sentence grounding on Charades-STA dataset under  official train/test splits.}
     \vspace{-11pt}
  \setlength{\tabcolsep}{1.5mm}{\begin{tabular}{c|c|cccc}
    \hline
     \multirow{2}*{Method} & \multirow{2}*{Type} & R@1, & R@1, & R@5, & R@5, \\ 
~ & ~ & IoU=0.5 & IoU=0.7 & IoU=0.5 & IoU=0.7\\
\hline
CTRL \cite{gao2017tall} & FS & 23.62 & 8.89 & 58.92 & 29.52\\
MMN \cite{wang2022negative}&FS&47.31& 27.28& 83.74& 58.41\\
2D-TAN \cite{zhang2019learning}& FS & 39.81 & 23.25 & 79.33 & 52.15\\
RaNet \cite{gao2021relation}&FS&43.87& 26.83& 86.67& 54.22\\
 DRN \cite{zeng2020dense} & FS & 45.40 & 26.40 & 88.01 & 55.38 \\
\hline
WSTAN \cite{wang2021weakly}& WS & 29.35 & 12.28 & 76.13 & 41.53 \\
ICVC \cite{chen2022explore}& WS & 31.02 & 16.53 & 77.53 & 41.91 \\
CNM \cite{zheng2022weakly}&WS&35.15&14.95&-&-\\
VCA \cite{wang2021visual}& WS & 38.13 & 19.57 & 78.75 & 37.75 \\
LCNet  \cite{yang2021local}& WS & 39.19 & 18.17 & 80.56 & 45.24 \\
    \hline \hline
    \textbf{Our TCSF} & \textbf{CD}& \textbf{53.85}&\textbf{37.20}&\textbf{90.86}&\textbf{58.95} \\ \hline
    \end{tabular}}
    \vspace{-4pt}
    \label{tab:Charades}
\end{table}
\begin{table}[t!]
\small
    \centering
    \caption{Performance comparison for temporal sentence grounding on  TACoS dataset under official train/test splits.}
     \vspace{-11pt}
    \setlength{\tabcolsep}{1.5mm}{\begin{tabular}{c|c|cccc}
    \hline
\multirow{2}*{Method} & \multirow{2}*{Type} & R@1, & R@1, & R@5, & R@5, \\ 
    ~ & ~ & IoU=0.3 & IoU=0.5 & IoU=0.3 & IoU=0.5  \\ \hline 
CTRL \cite{gao2017tall}&FS&18.32& 13.30& 36.69& 25.42\\
ACRN \cite{liu2018attentive}&FS &   19.52& 14.62&  34.97& 24.88\\
CMIN \cite{zhang2019cross}&FS &   24.64& 18.05&  38.46& 27.02\\
SCDM \cite{yuan2019semantic}&FS & 26.11& 21.17& 40.16& 32.18\\
 DRN \cite{zeng2020dense}&FS&-&23.17&-&33.36\\
2D-TAN \cite{zhang2019learning}&FS&37.29&25.32&57.81&45.04\\
MMN \cite{wang2022negative}&FS&39.24&26.17&62.03&47.39\\
FVMR \cite{gao2021fast}&FS& 41.48&29.12&64.53&50.00\\
RaNet \cite{gao2021relation}&FS&43.34& 33.54& 67.33& 55.09\\
MIGCN \cite{zhang2021multi}&FS&48.79& 37.57& 67.63& 57.91\\
    \hline \hline
    \textbf{Our TCSF} & \textbf{CD}& \textbf{49.82}&\textbf{38.53}&\textbf{68.60}&\textbf{59.89}\\ \hline
    \end{tabular}}
    \vspace{-5pt}
    \label{tab:TACoS}
\end{table}

\noindent \textbf{Implementation details.}
All the experiments are implemented by PyTorch with an NVIDIA Quadro RTX 6000.
For entropy decoding, following \cite{wu2018compressed,wang2019fast}, we use an  MPEG-4 decoder \cite{sikora1997mpeg} to decompress video bit-stream  for obtaining I-frame and P-frame.
As for query encoding, we embed each word to 300-dimension features by the Glove model \cite{pennington2014glove}. Besides, we set the head size of multi-head self-attention to 8, and the hidden dimension of Bi-GRU to 512, respectively. 
During training, we optimize parameter by  Adam optimizer with learning rate  $4\times 10^{-4}$ and linear learning rate decay of 10 for each 40 epochs. The batch size is 16 and the maximum training epoch is  100. We set $\alpha=0.8$ and $K=7$ in this paper.

\begin{table}[t!]
\small
    \centering
    \caption{Time complexity (s) of 100 videos on ActivityNet Captions dataset. The total time $T_{total}$ comprises the measurement time of decompressing video frames ($T_{dec}$), extracting the corresponding features ($T_{ext}$), and executing the network models ($T_{exe}$), where ``Other'' means the feature encoder (\textit{e.g.,} C3D/I3D).}
      \vspace{-11pt}
      \scalebox{0.95}{
    \setlength{\tabcolsep}{0.8mm}{
    \begin{tabular}{c|c|cccc|c|ccccccccccccc}
    \hline
    \multirow{2}*{Model} & \multirow{2}*{$T_{dec}$}&\multicolumn{4}{|c|}{$T_{ext}$} &\multirow{2}*{$T_{exe}$}& \multirow{2}*{$T_{total}$} \\ \cline{3-6}  
    ~ & ~&I-frame&MV&Residual& Other&~&~\\ \hline
        CTRL \cite{gao2017tall}&50.72&-&-&-&30.36&372.74&453.82\\
        RaNet \cite{gao2021relation}&50.72&-&-&-&30.36&406.30&487.38\\
    2D-TAN \cite{zhang2019learning}&50.72&-&-&-&30.36&434.91&515.99\\
        MIGCN \cite{zhang2021multi}&50.72&-&-&-&30.36&529.27&610.35\\
        MMN \cite{wang2022negative}&50.72&-&-&-&30.36&556.43&637.51\\
        DRN \cite{zeng2020dense}&50.72&-&-&-&30.36&585.72&666.80\\\hline
    TAG \cite{mithun2019weakly}&50.72&-&-&-&30.36&162.28&243.36\\
    WSTAN \cite{wang2021weakly}&50.72&-&-&-&30.36&183.86&264.94\\ 
    CNM \cite{zheng2022weakly}&50.72&-&-&-&43.86&175.37&269.95\\\hline \hline
    \textbf{Our TCSF}&\textbf{12.67}&\textbf{1.84}&\textbf{0.61}&\textbf{0.28}&-&\textbf{30.76} &\textbf{46.16}\\\hline
    \end{tabular}}}
    \vspace{-10pt}
    \label{tab:efficiency}
\end{table}

\subsection{Comparison with State-of-the-Arts}
We conduct performance comparison on three datasets. To evaluate efficiency, we only choose the open-source compared methods  that are grouped into two categories: (i) Fully-supervised (FS) setting  \cite{gao2017tall,liu2018attentive,yuan2019semantic,zhang2019cross,zhang2019learning,zeng2020dense,gao2021fast,zhang2021multi,gao2021relation,wang2022negative};
(ii) Weakly-supervised (WS) setting 
\cite{chen2022explore,yang2021local,zhang2020counterfactual,wang2021visual,wang2021weakly,zheng2022weakly}.
For convenience, we denote ``compressed-domain setting'' as ``CD''.
Following \cite{zhang2021natural,munro2020multi}, we directly cite the results of compared methods from corresponding works.  Note that no weakly-supervised  method reports its results on TACoS. The best results are \textbf{bold}.
From Tables~\ref{tab:ActivityNet}, \ref{tab:Charades} and \ref{tab:TACoS}, we can find that our  TCSF outperforms all compared methods by a large margin. It demonstrates that our model can achieve effective performance in more challenging compressed-domain setting.

\noindent \textbf{Efficiency comparison.} 
To fairly evaluate the efficiency of our TCSF, we conduct comparison on ActivityNet Captions dataset with some state-of-the-art methods whose source codes are available.
Table~\ref{tab:efficiency} reports the results, and we consider the decompressing time $T_{dec}$, the feature extracting time $T_{ext}$, the network executing time $T_{exe}$, where the time is measured via an average on the whole videos. As depicted in Table~\ref{tab:efficiency}, we have the following observations: (i) Our model takes 12.67s to decompress GOPs in each video bit-streams and 1.84s, 0.61s, 0.28s to extract their three features, which is much efficient than previous works. The main reason is that previous works need to decompress full frames of the video and rely on the heavy-weight 3D encoder like C3D/I3D to extract the features.
Instead, we need less frame-level context with much light-weight encoder.
(ii) Our network executing $T_{exe}$ also has less parameters to learn than previous work, thus achieving faster speed.
Overall, experimental results demonstrate the time-efficiency of our method.

\begin{table}[t!]
\small
\caption{Main ablation study on ActivityNet Captions dataset, where we remove each key individual component to investigate its effectiveness. ``PFG'' denotes ``pseudo feature generation'', ``TTA'' denotes ``Three-branch spatial-temporal attention'', ``AMF'' denotes ``adaptive motion-appearance fusion''.}
\vspace{-11pt}
\scalebox{1.0}{
\setlength{\tabcolsep}{1.5mm}{
\begin{tabular}{ccc|cccccccccccccccccc}
\hline
\multirow{2}*{PFG}&\multirow{2}*{TTA}&\multirow{2}*{AMF}& R@1 & R@1 & R@5 & R@5\\
~&~&~  & IoU=0.3 & IoU=0.5 & IoU=0.3 & IoU=0.5\\\hline
\XSolidBrush & \XSolidBrush & \XSolidBrush &  50.59&32.84&76.12&68.33\\
\CheckmarkBold & \XSolidBrush & \XSolidBrush &  60.25&41.82&79.10&72.08\\
\XSolidBrush & \CheckmarkBold & \XSolidBrush & 62.79&45.87&79.45&76.13 \\
\XSolidBrush & \XSolidBrush & \CheckmarkBold & 63.74&45.39&80.16&76.05  \\
\CheckmarkBold & \CheckmarkBold & \XSolidBrush & 64.19&47.56&83.77&76.90 \\\hline
\CheckmarkBold &\CheckmarkBold &\CheckmarkBold &\textbf{66.87}&\textbf{48.38}&\textbf{88.75}&\textbf{80.24}
\\ \hline
\end{tabular}}}
\vspace{-4pt}
\label{tab:ablation1}
\end{table}

\begin{table}[t!]
\small
\caption{Ablation study on pseudo feature generation.}
\vspace{-11pt}
\scalebox{1.0}{
\setlength{\tabcolsep}{1.3mm}{
\begin{tabular}{cc|ccccccccccccccccccc}
\hline
Appearance&Motion& R@1 & R@1 & R@5 & R@5\\
feature&feature& IoU=0.3 & IoU=0.5 & IoU=0.3 & IoU=0.5\\\hline
\XSolidBrush &\CheckmarkBold &64.73&47.51&88.09&78.10 \\
\CheckmarkBold & \XSolidBrush & 65.85&48.02&87.80&79.03\\\hline
\CheckmarkBold &\CheckmarkBold  &\textbf{66.87}&\textbf{48.38}&\textbf{88.75}&\textbf{80.24}
\\ \hline
\end{tabular}}}
\vspace{-4pt}
\label{tab:ablation_rfg}
\end{table}

\begin{table}[t!]
\small
\caption{Ablation study on three-branch spatial-temporal attention.}
\vspace{-11pt}
\scalebox{1.0}{
\setlength{\tabcolsep}{1.3mm}{
\begin{tabular}{cc|ccccccccccccccccccc}
\hline
Spatial&Temporal& R@1 & R@1 & R@5 & R@5\\
attention&attention& IoU=0.3 & IoU=0.5 & IoU=0.3 & IoU=0.5\\\hline
\XSolidBrush &\CheckmarkBold &64.56&43.82&84.13&77.50\\
\CheckmarkBold & \XSolidBrush & 65.31& 43.20&83.72&76.81\\\hline
\CheckmarkBold &\CheckmarkBold  &\textbf{66.87}&\textbf{48.38}&\textbf{88.75}&\textbf{80.24}
\\ \hline
\end{tabular}}}
\vspace{-4pt}
\label{tab:ablation_tta}
\end{table}

\begin{table}[t!]
\small
\caption{Ablation study on adaptive motion-appearance fusion.}
\vspace{-11pt}
\scalebox{0.9}{
\setlength{\tabcolsep}{0.8mm}{
\begin{tabular}{cc|ccccccccccccccccccc}
\hline
Query-guided&Residual-guided& R@1 & R@1 & R@5 & R@5\\
enhancement&fusion& IoU=0.3 & IoU=0.5 & IoU=0.3 & IoU=0.5\\\hline
\XSolidBrush &\CheckmarkBold &65.94&47.46&86.52&78.88\\
\CheckmarkBold & \XSolidBrush & 65.80&47.92&87.63&79.15\\\hline
\CheckmarkBold &\CheckmarkBold  &\textbf{66.87}&\textbf{48.38}&\textbf{88.75}&\textbf{80.24}
\\ \hline
\end{tabular}}}
   \vspace{-5pt}
\label{tab:ablation_amf}
\end{table}

\begin{figure*}[t!]
\centering
\includegraphics[width=\textwidth]{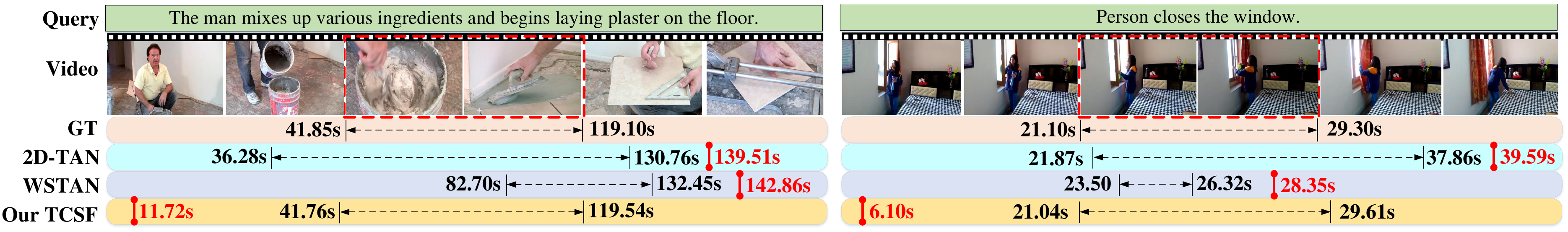}
\vspace{-18pt}
\caption{Qualitative prediction examples, where we complete the prediction at the \textcolor{red}{red} time. We find that our TCSF can ground earlier than the ground-truth start timestamp, while other methods ground later than the end timestamp.}
\label{fig:qualitative}
\vspace{-5pt}
\end{figure*}

\subsection{Ablation study}
To validate the effectiveness of each component in our TCSF, we conduct extensive ablation studies on the most challenging  ActivityNet Captions dataset.

\noindent \textbf{Main ablation studies.}
To analyze how each component contributes to the challenging task, we perform main ablation study as shown in Table~\ref{tab:ablation1}. Firstly, we set a baseline model that does not utilize pseudo feature, three-branch spatial-temporal attention module and adaptive motion-appearance fusion strategy to address the compressed-domain TSG. Similar to previous supervised methods, the baseline model directly generates multiple coarse segment proposals and then utilizes the rank loss for training. We can find that this baseline performs worse than most state-of-the-art methods in Table \ref{tab:ActivityNet}. Secondly, by designing the pseudo feature generation (PFG) module, we can effectively improve the performance since it enriches the full-frame context of the video. Table \ref{tab:ablation_rfg} further analyzes the effective of both pseudo appearance and motion features.
Thirdly, applying three-branch spatial-temporal attention (TTA) module also brings the large improvement since our well-designed spatial-temporal attention extracts the more fine-grained region-attentive temporal-spatial information for modelling more accurate activity content. As shown in Table \ref{tab:ablation_tta}, we further illustrate the effectiveness of spatial and temporal attentions separately.
Besides, the adaptive motion-appearance fusion (AMF) strategy also boost the performance a lot because it can balance the importance between appearance and motion features. Table \ref{tab:ablation_amf} illustrates the contributions of the query-guided feature enhancement and residual-guided fusion in AMF module. Overall, each component brings the performance improvement, and the full TCSF achieves the best results.


\begin{table}[t!]
\small
   \centering
    \caption{Effect of different low-level features.}
    \vspace{-11pt}
    \setlength{\tabcolsep}{1.1mm}{
    \begin{tabular}{ccc|cccccccccccc}
\hline
\multirow{2}*{I-frame}& \multirow{2}*{MV}& \multirow{2}*{Residual}& R@1,   & R@1,   & R@5, & R@5, \\
&&&IoU=0.3&IoU=0.5&IoU=0.3&IoU=0.5\\\hline
\CheckmarkBold & \XSolidBrush & \XSolidBrush & 64.27&46.83&85.75&76.54\\
\CheckmarkBold & \XSolidBrush & \CheckmarkBold & 65.66& 47.82&86.36&78.59\\
\CheckmarkBold & \CheckmarkBold & \XSolidBrush & 66.03&47.94&88.03&79.28\\\hline
\CheckmarkBold &\CheckmarkBold &\CheckmarkBold &\textbf{66.87}&\textbf{48.38}&\textbf{88.75}&\textbf{80.24}\\
\hline
   \end{tabular}}
   \vspace{-4pt}
    \label{tab:channel}
\end{table}
\noindent \textbf{Effect of different  low-level features.} To analyze the contribution of different low-level features, we conduct the ablation study as shown in Table \ref{tab:channel}. Both MV and residual can significantly improve the performance. The improvement shows the effectiveness of MV and residual.

\begin{table}[t!]
\small
    \centering
    \caption{Effect of the nouns-formed query in spatial attention.}
    \vspace{-11pt}
    \setlength{\tabcolsep}{2.8mm}{
    \begin{tabular}{c|ccccc}
    \hline 
 \multirow{2}*{Changes} & R@1, & R@1, & R@5, & R@5,  \\
~ & IoU=0.3 & IoU=0.5 & IoU=0.3 & IoU=0.5 \\ \hline
      w/o  query & 64.26&47.82&88.17&77.54 \\ 
 w/  query &\textbf{66.87}&\textbf{48.38}&\textbf{88.75}&\textbf{80.24}  \\ \hline
    \end{tabular}}
       \vspace{-5pt}
    \label{tab:noun}
\end{table}
\noindent \textbf{Effect of the nouns-formed query.}  In our spatial attention module, we utilize the noun feature to help us extract the spatial information. As shown in Table \ref{tab:noun}, we analyze the effect of the specific nouns-formed query. Based on the query, our TCSF improves the performance by 2.61\% in ``R@1, IoU=0.3''. This is because the nouns-formed query can locate the specific region for each frame, which reduces the  distraction of background information in the video.

\noindent \textbf{Analysis on the hyper-parameters.}
Moreover, we investigate the robustness of the proposed model to different hyper-parameters in Table~\ref{tab:ablation4}. In the temporal attention module, we  choose consecutive $K$ frame to extract the temporal information. 
We find we can obtain the best performance when $K=7$. In the grounding head module, we leverage $\alpha$ to balance the two losses. When $\alpha=0.8$, our TCSF obtains the best performance.

\subsection{Qualitative Results}
As shown in Figure~\ref{fig:qualitative}, we report the representative visualization of the grounding performance. Our  TCSF can  ground more accurate query-related segment boundaries than 2D-TAN and WSTAN with faster grounding.

\begin{table}[t!]
\small
\caption{Effect of different hyper-parameters.}
\vspace{-11pt}
\scalebox{1.0}{
\setlength{\tabcolsep}{1.2mm}{
\begin{tabular}{cc|cccccc}
\hline
\multirow{2}*{Module} &\multirow{2}*{Changes} & R@1 & R@1 & R@5 & R@5\\
&~ & IoU=0.3 & IoU=0.5 & IoU=0.3 & IoU=0.5\\
\hline
\multirow{3}*{\tabincell{c}{Temporal \\ attention}}&$K=6$& 66.28&47.95&87.34&80.17\\
~&\textbf{$K=7$}& \textbf{66.87}&48.38&\textbf{88.75}&\textbf{80.24}\\
~&$K=8$&65.92&\textbf{48.53}&86.11&79.30\\\hline
\multirow{3}*{\tabincell{c}{Grounding \\ head}} &$\alpha=0.7$ & 66.02 & 46.98 & 87.94 & \textbf{80.31} \\
~&\textbf{$\alpha=0.8$} &\textbf{66.87}&\textbf{48.38}&\textbf{88.75}&80.24 \\
~&$\alpha=0.9$ & 65.93&47.06&87.29&79.23
\\ \hline
\end{tabular}}}
   \vspace{-5pt}
\label{tab:ablation4}
\end{table}

\section{Conclusion}
In this paper, we introduce a brand-new compressed-domain setting into the temporal sentence grounding task to directly utilize the compressed video rather than decompressed frames.  To handle the challenging setting, we propose a novel Three-branch Compressed-domain Spatial-temporal Fusion (TCSF) framework to extract and aggregate three kinds of low-level visual features for  grounding. 
Experimental results on three challenging datasets (ActivityNet Captions, Charades-STA and TACoS) demonstrate that our TCSF significantly outperforms existing fully- and weakly-supervised methods.


\bibliographystyle{ieee_fullname}

\bibliography{egbib}

\end{document}